\documentclass[fleqn,10pt]{wlscirep}
\usepackage[utf8]{inputenc}
\usepackage[T1]{fontenc}
\usepackage{fancyhdr}
\usepackage{datetime}
\usepackage{multirow}
\usepackage{makecell}
\usepackage{amsmath,amssymb}
\usepackage{array}
\usepackage{hyperref}
\usepackage{graphicx} 
\usepackage{multirow}
\usepackage[table]{xcolor}
\usepackage{booktabs}
\usepackage{amsmath}
\usepackage[table,xcdraw]{xcolor}
\usepackage{colortbl}
\usepackage{comment}
\usepackage{wrapfig}
\usepackage{caption}
\usepackage{xr}
\usepackage[textsize=footnotesize]{todonotes}
\title{{Decision-oriented benchmarking to transform AI weather forecast access: Application to the Indian monsoon}} 

\author[1]{Rajat Masiwal}
\author[2,$\dagger$]{Colin Aitken}
\author[1,$\dagger$]{Adam Marchakitus}
\author[3]{Mayank Gupta}
\author[4]{Katherine Kowal}
\author[4,5]{Hamid A. Pahlavan}
\author[6]{Tyler Yang}
\author[1]{Y. Qiang Sun}
\author[2,7, 8]{Michael Kremer}
\author[8,*]{Amir Jina}
\author[6,9,*]{William R. Boos}
\author[1,4,*]{Pedram Hassanzadeh}

\affil[1]{Department of the Geophysical Sciences, University of Chicago, IL, 60637}
\affil[2]{Development Innovation Lab, University of Chicago, IL, 60637}
\affil[3]{Development Innovation Lab India, University of Chicago Trust, India, 560025}
\affil[4]{Data Science Institute, University of Chicago, IL, 60637}
\affil[5]{NorthWest Research Associates, Boulder, CO, 80301}
\affil[6]{Department of Earth and Planetary Science, University of California, Berkeley,  Berkeley, California, 94720}
\affil[7]{Kenneth C. Griffin Department of Economics, University of Chicago, IL, 60637}
\affil[8]{Harris School of Public Policy, University of Chicago, IL, 60637}
\affil[9]{Climate and Ecosystem Sciences Division, Lawrence Berkeley National Laboratory, Berkeley, California, 94720}
\affil[*]{Corresponding authors: pedramh@uchicago.edu, william.boos@berkeley.edu, amirjina@uchicago.edu} 
\affil[$\dagger$]{Equal contributions}

\begin{abstract}
Artificial intelligence weather prediction (AIWP) models now often outperform traditional physics-based models on common metrics while requiring orders-of-magnitude less computing resources and time~\cite{pathak2022fourcastnet,bi2023accurate,lam2023learning,price2025probabilistic,ben2024rise,allen2025end}. Open-access AIWP models thus hold promise as transformational tools for helping low- and middle-income populations make decisions in the face of high-impact weather shocks~\cite{hallegatte2012cost,WMO2019AgriFoodSecurity,rosenzweig2019assessing,burlig2024long}. Yet, current approaches to evaluating AIWP models~\cite{rasp2024weatherbench,nathaniel2024chaosbench, nguyen2025indiaweatherbench} focus mainly on aggregated meteorological metrics without considering local stakeholders' needs in decision-oriented, operational frameworks. Here, we introduce such a framework that connects meteorology, AI, and social sciences. As an example, we apply it to the 150-year-old problem of Indian monsoon forecasting~\cite{blanford1884ii,gadgil2003indian}, focusing on benefits to rain-fed agriculture, which is highly susceptible to climate change\cite{hultgren2025impacts}. AIWP models skillfully predict an agriculturally relevant onset index at regional scales weeks in advance when evaluated out-of-sample using deterministic and probabilistic metrics. This framework informed a government-led effort in 2025 to send 38 million Indian farmers AI-based monsoon onset forecasts~\cite{vallangimonsoon25}, which captured an unusual weeks-long pause in monsoon progression. This decision-oriented benchmarking framework provides a key component of a blueprint for harnessing the power of AIWP models to help large vulnerable populations adapt to weather shocks in the face of climate variability and change~\cite{goswami2006increasing,wang2015rethinking}.

\end{abstract}
\begin{document}

\flushbottom
\maketitle
\thispagestyle{empty}

\section*{Introduction}

Accurate weather forecasts, tailored to the needs of local stakeholders, are critical for decision-making in many applications. Climate change increases this need for reliable forecasts, as changing weather conditions make historical information potentially less useful\cite{wang2015rethinking, katzenberger2020robust}, making forecasts key tools for climate adaptation. Despite the significant socioeconomic importance of high-quality forecasts\cite{hallegatte2012cost,WMO2019AgriFoodSecurity,rosenzweig2019assessing}, they are inaccessible to many people, due to, for example, low quality of some forecasts, technical and economic barriers to producing forecasts, or lack of investment in forecast dissemination tailored to users' decisions \cite{kull2021value, linsenmeier2023global}. Many lacking access live in low- and middle-income countries (LMICs) and were largely left behind during the first revolution of weather forecasting, which saw the development of increasingly sophisticated numerical weather prediction (NWP) models that require high-performance computing resources to run \cite{bauer2015quiet,linsenmeier2023global}. In some cases, this has resulted in one-day-ahead forecasts in low-income countries performing worse than seven-day-ahead forecasts in high-income countries\cite{linsenmeier2023global}. 

Neural network-based AIWP models are driving a second revolution in weather forecasting\cite{pathak2022fourcastnet,bi2023accurate,lam2023learning}, with the potential to greatly expand global access to high-quality forecasts. Open-source instances of some of these models offer not only higher accuracy and orders-of-magnitude faster computation than NWP models, as is often highlighted\cite{lang2024aifs,fuxi2023,lam2023learning,kochkov2024neural}, but also have the potential to transform the ability of LMICs to autonomously generate multi-model, ensemble forecasts. Moreover, the rapid development, generation, and evaluation of forecasts can enable production of portfolios of forecasts tailored to stakeholders' needs. 
This timely opportunity to democratize access to forecasts, coinciding with efforts to increase public availability of real-time weather data\cite{ecmwf2025opendata}, has generated activity by governments, philanthropies, the AI industry, academia, and others seeking to shift the paradigm of impact-based early-warning efforts, especially for LMICs, many of which are in the tropics~\cite{afs_wmo_funding_news,wmo_call,linsenmeier2023global}.

Benchmarking---quantitatively comparing the performance of models with one another and an appropriate baseline---is essential for this transformation, and has played a key role in advancing AI and its applications\cite{krizhevsky2012imagenet,jumper2021highly}. However, developing benchmarks that reliably inform choices about the adoption and large-scale dissemination of AI-based forecasts in real-time is a complex task. For such ``decision-oriented'' benchmarking, it is essential not only to consider meteorological perspectives, but operational constraints and societal needs. 
However,  traditional forecast verification practices often do not consider the latter, e.g., instead targeting large-scale dynamical metrics such as upper-level circulation (e.g., 500~hPa geopotential height) and extratropical regions where most NWP models were developed~\cite{buizza2015forecast}. Additionally, in many event-oriented benchmarking efforts, hindcasts are initialized before an event's known occurrence time, making it difficult to test for ``false alarms'' and ``false negatives'' that could occur often in real-time operational forecasts~\cite{alessandri2015prediction}.

Benchmarking is especially important for AIWP models given their black-box, data-driven nature, which can lead to a lack of known physical constraints\cite{bonavita2024some,sun2025can}, and their use of most available high-quality historical data for training, which results in a small dataset left for \textit{testing}. Rapidly emerging AIWP benchmarks~\cite{rasp2024weatherbench,nathaniel2024chaosbench, gupta2025observations,nguyen2025indiaweatherbench} provide valuable information on forecast accuracy from a dynamical meteorological perspective, but are not designed to inform strategic choices regarding operational deployment and large-scale dissemination to end users. Such choices could include agricultural agencies selecting a forecasting system to inform farmer planting decisions, public health agencies designing a system for heat warnings, or disaster managers setting risk levels to activate evacuation procedures. Thus, there is an unmet need for decision-oriented, operational benchmarking of AIWP models that is also grounded in climate science and AI. 

\begin{figure}[ht!]
  \centering
    \includegraphics[scale=0.70]{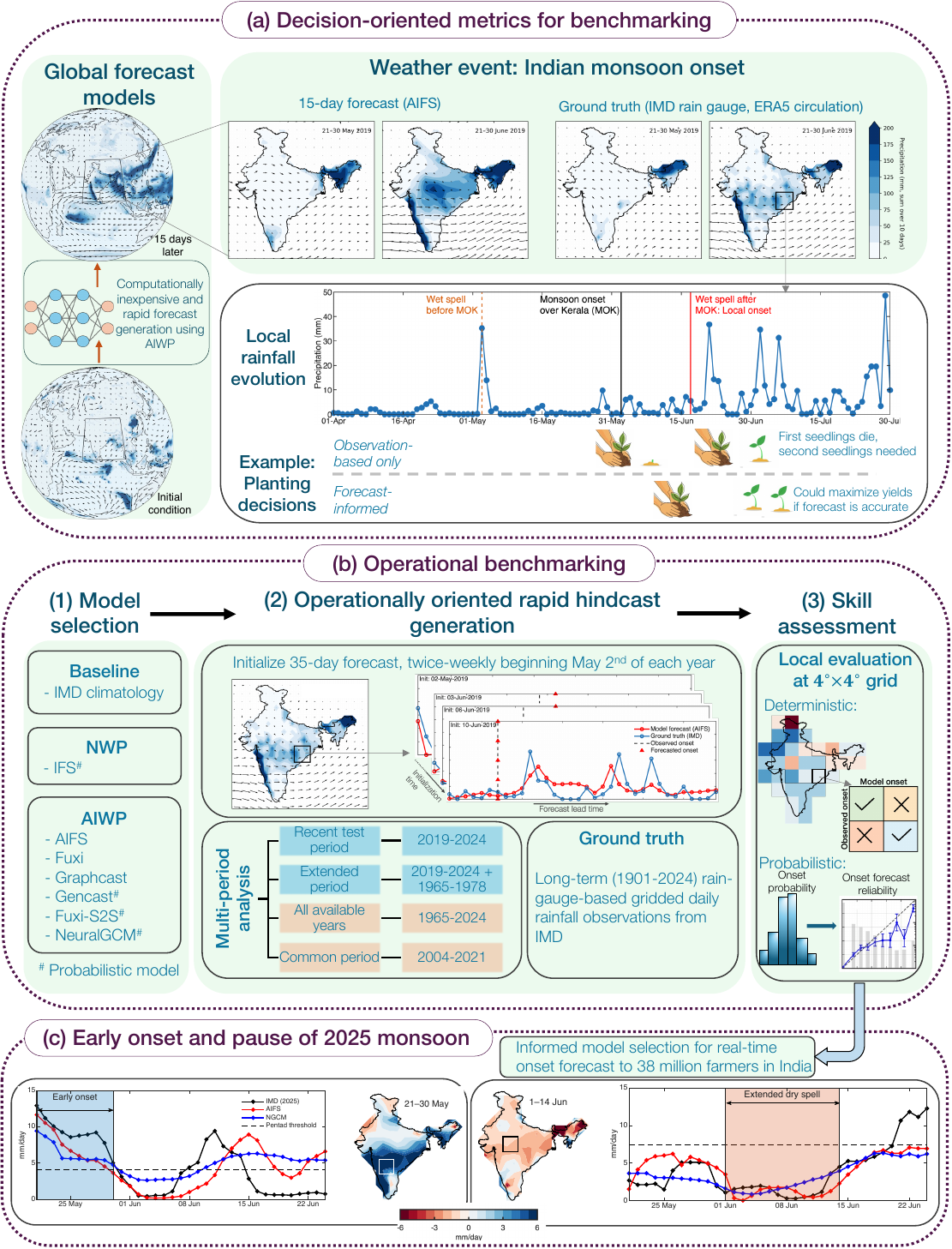}
  \caption{\textbf{Decision-oriented operational benchmarking framework: the Indian monsoon onset example}. Panels (a)–(b) present an operationally oriented benchmarking framework for global AIWP models using locally relevant, decision-oriented metrics. One example is an agriculturally relevant metric of the Indian summer monsoon onset, which marks the beginning of the primary rainy season over India. As shown in (a), the timing of the onset (marked by a red line), followed by continuous rainfall, is relevant to planting decisions and, consequently, for achieving good crop yields. Panel (b) outlines the key steps involved in the operational benchmarking of monsoon onset forecasts. 35-day forecasts from a suite of NWP and AIWP models are generated or retrieved for analysis. To expose the forecasts to false alarms and onset misses, in an operationally oriented setting, twice-weekly initializations from early May are used, and the forecasts are evaluated using both deterministic and probabilistic metrics against IMD gridded rain-gauge observations (rather than ERA5, which was used for training). This framework assesses model skill in capturing local monsoon onset over $\approx$ 400 km regions across India. The analysis is conducted across multiple time periods, both including and excluding the AI models’ training years, to address the {\it small test sample size}. This framework informed model selection for a large-scale dissemination of AI-based forecasts at the medium-range and subseasonal timescale to 38 million farmers in 2025~\cite{vallangimonsoon25}. Panel (c) shows that the AI forecasts captured the early onset and the pause in the monsoon progression. Multi-model, probabilistic forecasts based on these models~\cite{aitken2026monsoonhybrid} provided timely warnings to farmers.}
  \label{fig1}
\end{figure}

Here, using tools and insight from climate science, AI, and development economics, we introduce a framework for decision-oriented, operational benchmarking that can inform large-scale dissemination. We demonstrate and test this framework for a high-impact target: supporting planting decisions of low-income farmers in India, where much agriculture relies on a single $\sim$4-month rainy season, and where extended-range forecasts of monsoon rains have remained an elusive goal for at least 150 years\cite{blanford1884ii,normand1953monsoon, gadgil2003indian,shukla2007monsoon}. In this work, we evaluate forecasts of the local monsoon onset date by six state-of-the-art global AIWP models, chosen because they forecast precipitation and are open-source or provide public data (Figure~\ref{fig1}). The local monsoon onset date, on which continuous rains start in a particular region, is an especially important prediction for agriculture~\cite{rosenzweig1993wealth,gadgil2006indian} (Figure~\ref{fig1}). While evidence shows that farmers can act to benefit from forecasts of local onset issued weeks in advance~\cite{burlig2024long}, such localized forecasts have not yet been systematically benchmarked or widely operationalized. Most publicly available forecasts have targeted only monsoon onset over Kerala (MOK) at the southern tip of India\cite{sarkar2025operational}. 

Skillfully forecasting local monsoon onset weeks in advance is expected to be challenging, given the difficulty of predicting precipitation at long lead times~\cite{shrestha2013evaluation,surcel2015study}. Additionally, AIWP models are typically trained to minimize global loss functions (e.g., Eqs.~\eqref{eq:loss}-\eqref{eq:inf}) on sub-daily time scales, and thus have not been optimized to capture this particular atmospheric regime transition. Furthermore, most AIWP models are trained on ERA5 precipitation, which is known to have shortcomings (see Methods). Yet, this benchmark identified skillful forecasts and informed AIWP model selection for a project led by India's Ministry of Agriculture and Farmers Welfare\cite{vallangimonsoon25}. In this project, two AIWP models were used in a blending algorithm that produced farmer-centric, probabilistic local monsoon onset forecasts for 38 million farmers in 2025\cite{aitken2026monsoonhybrid}.

\section*{Framework for decision-oriented forecast benchmarking}

Benchmarking to support adoption and dissemination requires a systematic event definition (e.g., an index), data for validation (``ground-truth''), and a reference against which skill will be assessed (the ``baseline''). For this decision-oriented benchmarking, we devise an agriculturally targeted index for rainy-season onset that is compatible with operational forecasting constraints and relevant for planting decisions.  Our index is also local in nature, enabling assessment of spatial variations in skill that can inform model selection and forecasting system design for large-scale dissemination. This differs from onset indices that are based on either the continental-scale transition of monsoon winds\cite{webster1992monsoon,pai2009summer} or universal rainfall thresholds inappropriate for local climates to which agricultural practices have adapted~\cite{sivakumar1988predicting,marteau2011onset}.

We modify a widely-used agronomic index \cite{moron2014interannual}, employed in past studies of farmer response to forecasts\cite{burlig2024long}, that defines local onset as the start of the first 5-day rainy period that is typical for a location's wet season 
(Figure~\ref{fig1}(a); see Methods). That original index also requires a dry period to not follow within 30 days to discriminate between false onsets and the start of continuous rains. Because implementing that criterion explicitly would require forecasts for an additional 30 days beyond the lead time, we modify it with a constraint based on the median climatological declared date of MOK by the India Meteorological Department (IMD; see Methods).

\begin{figure}[ht!]
  \centering
  \includegraphics[width=0.8\textwidth]{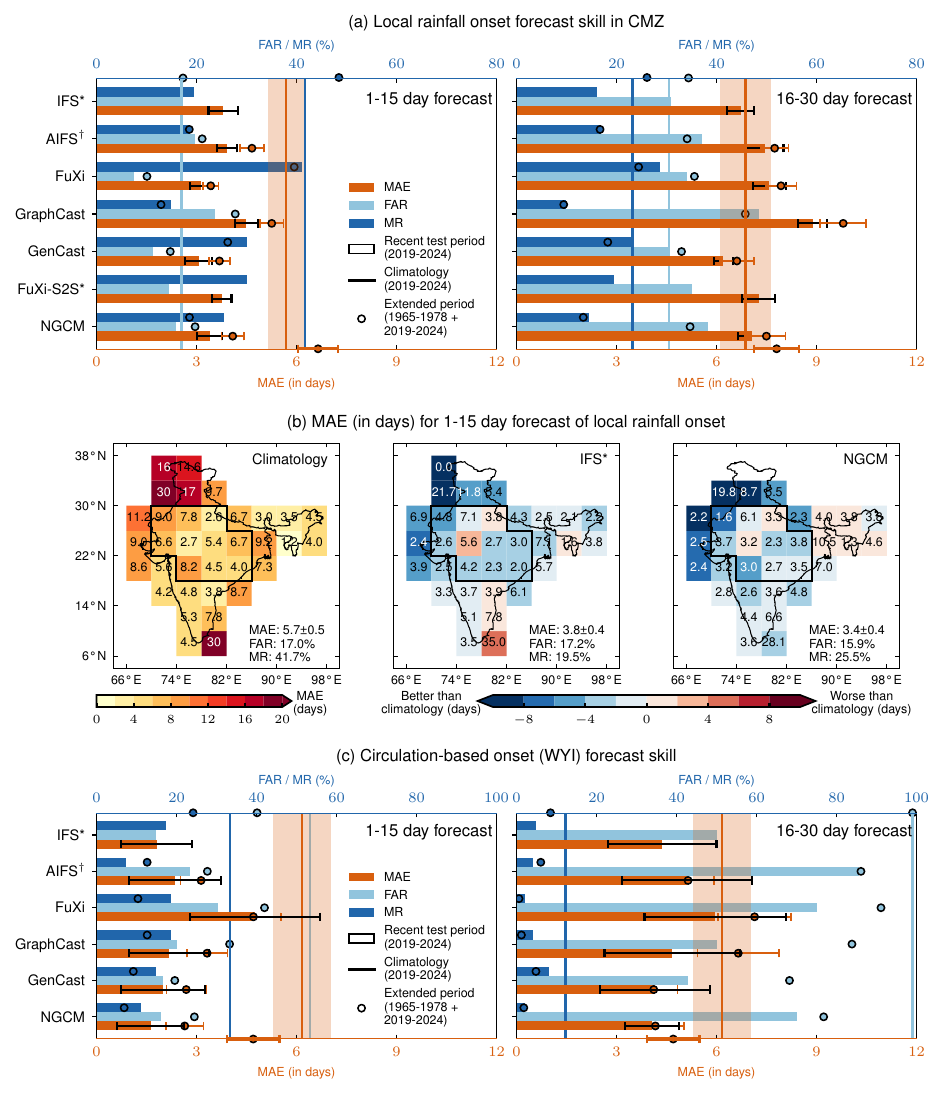}
  \caption{\textbf{Deterministic forecast skill of the models in the medium-range (1-15 day) and subseasonal (16-30 day) timescales across two analysis periods.} (a) Forecast skill of local rainfall-based monsoon onset (defined in Methods), averaged over the CMZ (outlined in panel (b)) at 1-15 day (left) and 16-30 day (right) lead times. Horizontal bars and vertical lines show model forecasts and climatological baseline, respectively, for the recent test period (2019-2024). Error bars and shading represent the standard error for the forecasts’ and climatology’s MAE. Models with $*$ next to them have some years unavailable in this period; models with $\dagger$ include some of their training or fine-tuning years within this period (see Extended Data Table~1). Circles show results for the extended period (1965–1978; 2019–2024), with climatological baseline scores placed on the horizontal axes. (b) The 1–15 day forecast MAE (in days) of climatology, IFS, and NGCM for the recent test period is shown in each $4^\circ \times 4^\circ$  box. Shading shows the MAE difference between IFS/NGCM and climatology in the middle and right panels (blue indicates improved skill). Average MAE, FAR, and MR over the CMZ are also shown for each model. (c) Same as (a), but for a large-scale circulation-based monsoon onset defined using the Webster-Yang index (WYI)~\cite{webster1992monsoon} (see Methods). Spatial maps similar to (b) for a 16-30 day forecast window and averaged scores for the common model period (2004-2021) are presented in Extended Data Figure~2.}
  \label{fig2}
\end{figure}

We focus primarily on the core monsoon zone (CMZ, outlined in black in Figure~\ref{fig2}b), which constitutes a large fraction of all-India summer monsoon rainfall and exhibits coherent intraseasonal variability~\cite{rajeevan2010active}. This region  
includes states that had over 200 million agricultural workers in the most recent  census\cite{icrisat_dld_2011}. For assessing skill (i.e., accuracy relative to a baseline), we use historical average onset dates (a traditional climatological mean)  in 4$^\circ \times$4$^\circ$ grid cells across India (Figure~\ref{fig1}(b)). This simple reference forecast is a common baseline, especially at longer timescales (subseasonal-to-seasonal), and is often part of local knowledge.
For ground-truth, we employ IMD rain-gauge observational data\cite{rajeevan2006high} (see Methods). To address the small out-of-sample test data size (given that {\it one} onset occurs per year), we  produced hindcasts as far back as 1965 using the open-source models, evaluating them in four periods (Figure~\ref{fig1}). 

Benchmarking requires evaluation metrics, and for this decision-oriented process we include the commonly used mean absolute error (MAE) to measure the magnitude of timing errors, as well as the miss rate (MR) and false alarm rate (FAR).  The latter are expected to be important for planting decisions\cite{burlig2024long}; forecasts that guide farmers to plant much too early (a false alarm for onset; see Methods) risk crop failure when germination occurs long before sufficient rain. We also employ probabilistic metrics, because research has shown that farmers prefer probabilistic forecasts~\cite{murphy1977_value_of_forecasts, krzysztofowicz1983_bayes_forecasting, krzysztofowicz2001_probabilistic_hydrology} and AIWP models enable widespread generation of ensemble forecasts. The risk profile of end users will influence the weight given to each metric. For example, farmers with no savings might  especially penalize high FAR or plant only when forecast probabilities are very high, as discussed in related work\cite{aitken2026monsoonhybrid}. 

\section*{Evaluating deterministic skill}
Of six AIWP models and one NWP model evaluated for the recent test period of 2019 to 2024 (Extended Data Table~1), nearly all provide skillful deterministic forecasts of this  onset index at medium-range (1-15 day) timescales (Figure~\ref{fig2}, Extended Data Figures~1 and 2). In the CMZ, most models on average outperform climatology in terms of MAE and MR at lead times of 1-15 days (Figure~\ref{fig2}a), while having comparable FAR (Supplementary Table S1). 

Examining the models spatially, we see that an NWP model widely regarded as having high skill in general (ECMWF's IFS) and an example AIWP model (Google's NeuralGCM, NGCM \cite{yuval2024ngcmimerg}) both outperform climatology for the 1-15 day forecast window in most regions (Figure~\ref{fig2}b).  Within the CMZ, most models' error is around two days smaller than that of the climatological forecast for 1-15 day lead times. The deterministic accuracy of most models approaches or falls below that of the baseline for lead times of 16-30 days (Figure~\ref{fig2}a), but some (IFS and GenCast) retain skill out to $\sim$3 weeks of lead time by all three deterministic metrics (Extended Data Figures~1 and 2b). Model skill can be compared across regions, lead times, and verification metrics (Supplementary Figures S1–S6). However, uncertainty due to factors such as small sample size precludes a definitive ranking of models. A robust finding is that no single model consistently outperforms all others across all metrics and regions. This suggests potential benefit of blending multiple models operationally, as done for the 2025 monsoon onset dissemination~\cite{aitken2026monsoonhybrid}. 

Since many AIWP models were trained on reanalyses covering 1979 to around 2018, and fine-tuned on more recent years (Extended Data Table 1), this might seem to leave at most $\sim$6 onset events for out-of-sample testing. Unlike most NWP models, however, AIWP models can be used to rapidly generate hindcast data with the identical model version that will be used operationally. This enables a more informative comparison than typical NWP hindcasts, whose configurations often differ from operational forecasting setups (with major implications for the widely used model averaging or blending process \cite{aitken2026monsoonhybrid}). India provides a unique testbed for evaluating AIWP performance for tropical applications because its weather records extend decades prior to the observational data used for AIWP training. Therefore, challenges arising from small test sample sizes can be alleviated by including earlier periods (e.g., before 1979), with the caveat that initial conditions may be less accurate in that pre-satellite period.  We  produced hindcasts for 1965-1978 for all the open-source AIWP models (Extended Data Table 1 and Supplementary Table S2), demonstrating how LMICs' meteorological agencies could employ these models to benchmark forecasts using their own national weather data and generate larger hindcast sets for events with small sample size (e.g., weather extremes).

Model skill appears higher for the extended period (1965-1978 + 2019-2024), because the climatological baseline performs worse for this period (Supplementary Figure S7).  Specifically, the baseline MAE increases by about a day, possibly because onset dates were less anomalous after 2019 (Figure~\ref{fig2}a, Extended Data Figure~2, Supplementary Table 3).  This improvement makes the 16-30 day lead-time forecasts at least as good as climatology for most models within this period, with GraphCast being the only model that has MAE and FAR values worse than baseline for all periods, and FuXi and FuXi-S2S suffering from total misses (no onset in a summer) despite their competitiveness in other metrics (Supplementary Table S1 and S3-5). In the extended period, GenCast stands out with its MAE, FAR, MR, and total misses comparable to or better than other models. However, generating these long hindcasts with GenCast, a diffusion-based model, incurs high computational cost and long runtime (Methods), hindering its inclusion in analyses of other periods (Supplementary Tables 4-5). Overall, these results demonstrate the importance of comparing with a meaningful baseline, and the utility of rapid hindcast generation by AIWP models for enabling such comparison in diverse test samples.

In addition to an increase in rain rate, monsoon onset also involves a transition in continental-scale winds\cite{webster1992monsoon,boos2009annual} which a dynamically consistent forecast should capture.
Using an index based on large-scale wind\cite{webster1992monsoon} to define the onset (see Methods), we find that all models outperform the baseline for a 15-day forecast (Figure~\ref{fig2}c). Most models have an average error that is 3-4 days smaller than that of the baseline at 1-15 day lead times. In contrast to the local rainfall metric, positive skill for this dynamical metric is preserved for most models at 16-30 day lead times (the FAR metric loses meaning at these extended lead times because the large-scale dynamical transition occurs every year with only modest variance in its timing\cite{webster1992monsoon}). 
The relatively higher skill in forecasting this ERA5-based circulation index may be due to a few factors, including the relatively higher predictability of the large-scale tropical circulation and data consistency, as most AIWP models are trained on ERA5 circulation data.

\section*{Evaluating probabilistic skill}

The computational efficiency (and, in some cases, stochastic architecture) of AIWP models enables the production of ensemble forecasts by countries lacking supercomputing resources. These ensembles, in turn, enable probabilistic forecasts that address initial-condition uncertainty and are especially attractive for applications such as agriculture, where stakeholders make critical decisions based on information to which forecasters do not have full access\cite{aitken2026monsoonhybrid}. Decision-theoretic work has shown that probabilistic forecasts are more valuable than deterministic forecasts in such settings \cite{murphy1977_value_of_forecasts, krzysztofowicz2001_probabilistic_hydrology}, motivating the benchmarking of raw forecast probabilities (i.e., fraction of ensemble members predicting an event) against observed likelihoods. 

\begin{figure}[t!]
  \centering
\includegraphics[width=\linewidth]{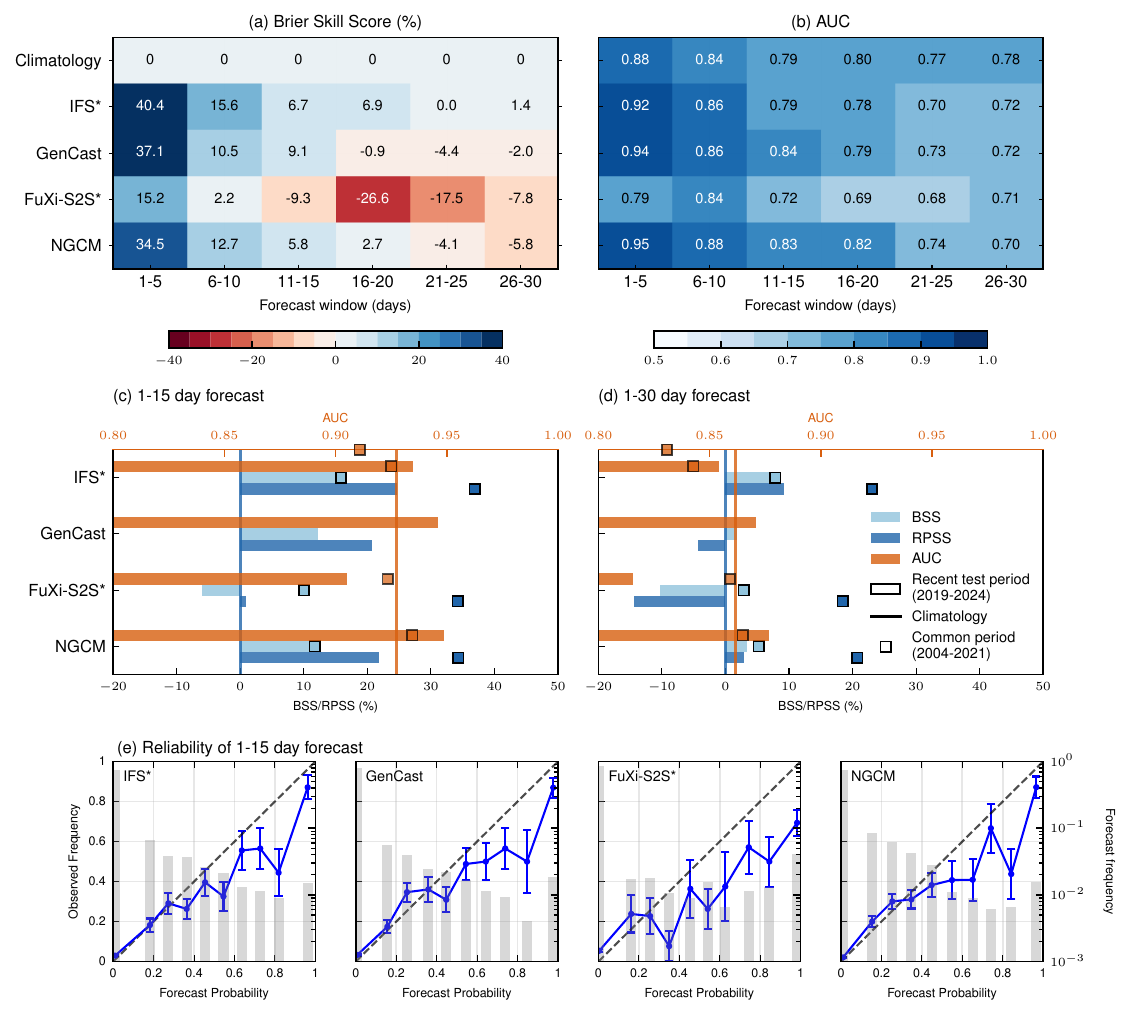}
  \caption{\textbf{Probabilistic performance of the models over the CMZ.} (a) Brier skill score (BSS) and (b) area under the receiver operating characteristic curve (AUC) for 5-day binned forecasts over the 10 grid cells of the CMZ for the recent test period (2019–2024) (see Methods for details). BSS is computed relative to a climatological forecast, with positive values indicating improved performance compared to climatology. For AUC, higher values mean better performance. Panels (c) and (d) show aggregated probabilistic scores for the 1–15- and 1–30-day forecasts, respectively. These scores are obtained by binning the onset forecasts into 5-day windows. To account for differing numbers of ensemble members, the fair BSS and ranked probability skill score (RPSS) are used (Methods). Climatological baselines for the respective metrics during this period are shown as vertical lines. Scores for the common period (2004–2021) are indicated by square markers, with climatological scores placed on the horizontal axes (GenCast's forecasts for this period were not produced due to computational constraints). (e) Reliability diagram for onset forecasts from four probabilistic models for the 1–15-day forecast window for the recent test period. Confidence intervals of two standard errors around the observed frequencies are shown. Models marked with $*$ have year(s) unavailable during this period. More Results from (a)-(e) but for the common period are presented in Extended Data Figure~3.}
  \label{fig3}
\end{figure}

Probabilistic metrics show skill for some models extending to three weeks of lead time (Figure~\ref{fig3}a, b). This was demonstrated for the 2019-2024 period, when onset dates were less anomalous and when it will thus be more difficult for models to outperform the baseline.  Indeed, the ranked probability score (lower values are better) of a {\it climatological forecast} in the recent test period is only about 0.5, compared to about 0.75 in the 2004-2021 common period shared by all models. The IFS ensemble and two of the three probabilistic AIWP models (GenCast and NGCM) have skillful 15-day forecasts by all three probabilistic metrics examined (Figure~\ref{fig3}c).  Aggregating across lead times from 1-30 days shows that IFS and NGCM have positive Brier Skill Scores (BSS); the IFS' BSS is higher, but its  area under the receiver operating characteristic curve (AUC) metric is lower than that of both NGCM and the baseline (Figure~\ref{fig3}d and Extended Data Figure 3d).   

Some models fare worse by probabilistic metrics: FuXi-S2S has a negative BSS for its 15-day forecasts, contrasting with its skillful deterministic metrics (MAE, FAR, and MR were all better than baseline at that lead time).  However, this model has only three years of out-of-sample test data available (it was the only non-public model, but with limited public forecasts, that we benchmarked) and performed better when evaluated during a period overlapping with its training set (2004-2021; Extended Data Figure 3).  This difference between probabilistic and deterministic skill for FuXi-S2S may be an artifact of the small sample size or may indicate correlations among its ensemble members. This case illustrates the importance of making model weights for inference publicly available to enable robust benchmarking.

The models are generally overconfident, with reliability diagrams showing predicted probabilities that are higher than observed frequencies for most forecasts (Figure~\ref{fig3}e).  These overconfident predictions might be at least partially remedied by probability calibration schemes \cite{gneiting2005weather,wilks2009extending}, which may be most desirable to do in a multi-model ensemble, as described in a related paper\cite{aitken2026monsoonhybrid}.

The above probabilistic results focused on the CMZ, and the models generally have higher probabilistic skill in this region and in areas immediate south and west (Extended Data Figure~4).  Probabilistic skill is lower north and east of the CMZ, in contrast to the higher deterministic skill in those regions (compare Extended Data Figure~4 with Figure~\ref{fig2}b). Additionally, the area in which the models outperform climatology is larger for the ranked probability score than for the Brier score, because the former rewards predictions that correlate with onset but do not land in the identical five-day bin as the observed onset.

\section*{Utility in the abnormal 2025 monsoon season}

\begin{figure}[htb!]
  \centering
\includegraphics[width=\linewidth]{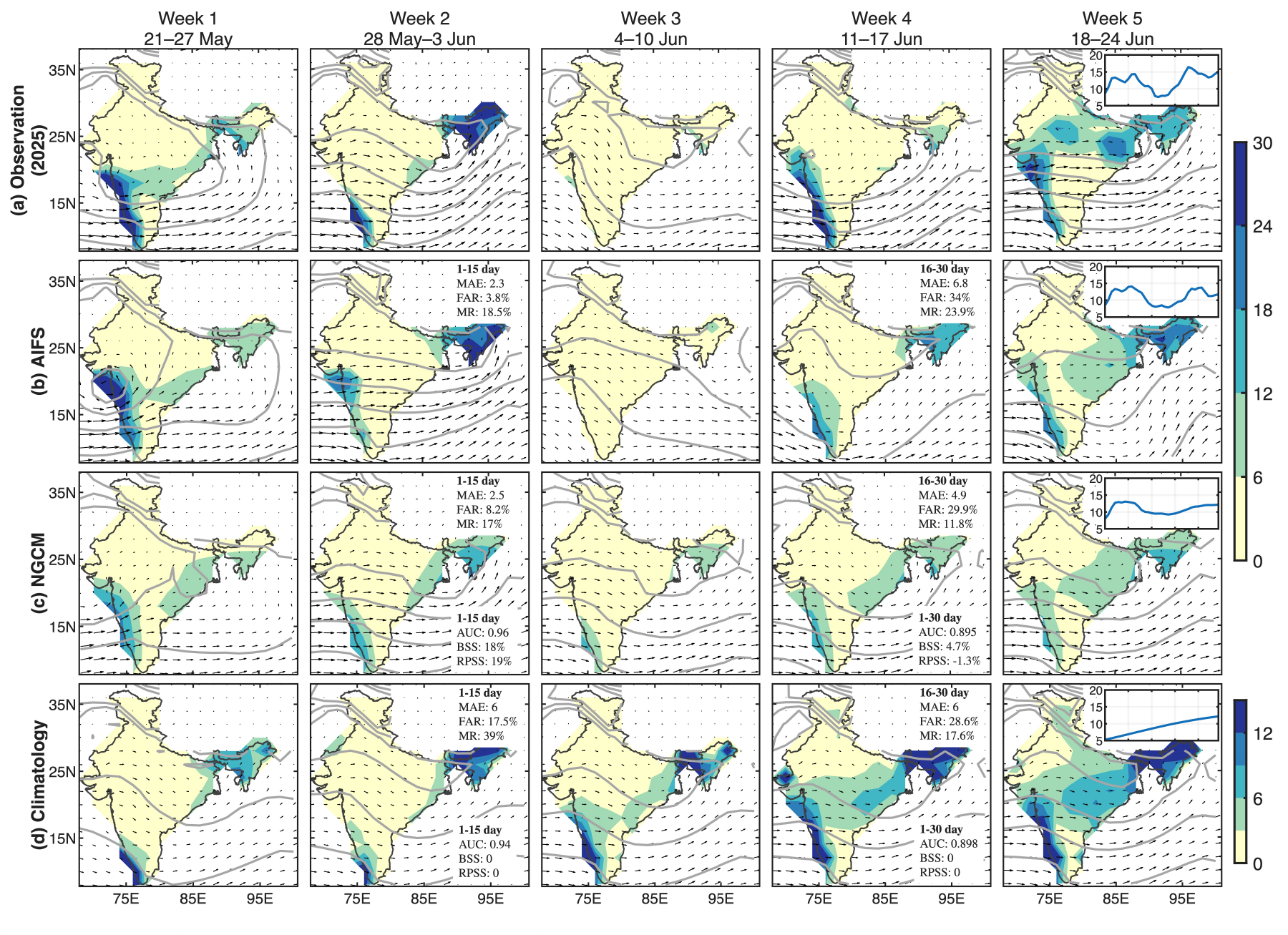}
\caption{\textbf{AIWP models capture the early onset followed by an extended dry spell during the 2025 Indian monsoon season.} Progression of weekly-averaged rainfall (shading, mm/day), 850~hPa winds (vectors), and 850~hPa geopotential (contours) for (a) IMD real-time data from 2025; (b)-(c) forecasts from AIFS and NGCM initialized on 20 May 2025, and (d) climatology. The insets in the last columns show the evolution of the Somali Jet index over five weeks. Both AIWP models qualitatively capture the anomalously early onset over southern India (week~1) and the subsequent pause in its northward progression, leading to a 2--3-week dry spell over central India (28 May--17 June). For NGCM, ensemble-mean rainfall and circulation are shown. Panels for week 2 and week 4 also include onset forecast skill metrics (see Extended Data Table 2 for more details). In (d), rainfall climatology is from IMD 1$^\circ$ data \cite{rajeevan2006high}, while wind and geopotential climatology are from ERA5. Rainfall in (a) is obtained from IMDLIB \cite{nandi2024imdlib}, which uses real-time IMD data at 0.25$^\circ$. All variables are regridded to a 2$^\circ$ resolution for comparison. AIWP models are initialized with ERA5; results for forecasts initialized with real-time analysis data are presented in Extended Data Figure~5.}
\label{fig4}
\end{figure}

The framework discussed above informed model selection for a recent large-scale forecast dissemination effort by India's Ministry of Agriculture and Farmers' Welfare. Based on the above results and practical considerations related to real-time operations (see Discussion), two AIWP models, AIFS and NGCM, were chosen for this project, which disseminated \textcolor{black}{calibrated probabilistic} onset forecasts via text messages to approximately 38 million farmers\cite{vallangimonsoon25}. The \textcolor{black}{model blending algorithm} and its performance are described in Aitken \emph{et al.}~\cite{aitken2026monsoonhybrid}. Here, we show that AIFS and NGCM individually captured the anomalous behavior of Indian monsoon onset in 2025, when a very early onset in southern India (MOK) was followed by over two weeks of dry conditions and a pause in the typical northward progression of local monsoon onset (Figures~\ref{fig1}c and \ref{fig4}a).  This anomalous behavior served as a high–societal-impact, out-of-sample test for AIWP models. 

Both AIFS and NGCM captured the overall spatio-temporal pattern of anomalous onset in 35-day forecasts initialized on 20 May 2025 (Figure \ref{fig4}b,c). Both models predicted the unusually heavy rainfall in late May in southern India associated with this early onset,  
then predicted the subsequent dry period of almost 3 weeks that covered much of central India (Figures \ref{fig1} and \ref{fig4}a). This unusual onset progression contrasts with the climatological behavior, where a systematic northward advance of monsoonal rainfall starts in early June (Figure \ref{fig4}d).

This qualitative agreement with observations is supported by quantitative improvements in forecast skill of the models relative to the baseline across most of the metrics in our benchmarking framework (Figure \ref{fig4} and Extended Data Table 2). 

These two models also successfully predicted the evolution of large-scale monsoon circulation weeks in advance, capturing the strengthening of low-level eastward flow across peninsular India and the Bay of Bengal at the two-week lead time and the subsequent weakening of that flow in the third week (Figure~\ref{fig4}).  The early burst of large-scale monsoon activity in late May, followed by a pause and reinvigoration in mid-June, can easily be seen in the intensity of the Somali jet, which constitutes the low-level inflow to the Indian monsoon (insets in Figure \ref{fig4}). Both AIFS and NGCM predicted this behavior with a lead time of 3-4 weeks, providing information not captured in the climatology of the Somali jet. 

To use AIWP models operationally, they must be initialized with real-time atmospheric state estimates (i.e., {\it analyses}) instead of the ERA5 {\it reanalyses} on which they were trained; this is necessary because the ERA5  data become available with delays of a few days. We find no substantial qualitative differences between forecasts initialized on May 20th with real-time analyses and with ERA5 (Extended Data Figure~5), although detailed analyses show some differences (Supplementary Table S6). 

\section*{Discussion}

Billions of people in LMICs lack access to reliable, tailored weather forecasts that can improve their livelihoods\cite{kull2021value,linsenmeier2023global}, particularly in a changing climate\cite{goswami2006increasing,wang2015rethinking,katzenberger2020robust}. Open-source AIWP models can help climate adaptation by enabling LMIC governments and other institutions to \textit{(i)} evaluate and improve forecasts using local data, population-relevant decision-oriented metrics, and techniques such as bias correction and fine-tuning, and \textit{(ii)} generate real-time, multi-model ensemble forecasts with far fewer resources than required for traditional NWP. The former is enabled by the ability to rapidly generate decades of hindcasts with the same model version used operationally, and by the open-source, technically accessible nature of many AIWP models. Moreover, \textit{(i)} and \textit{(ii)} can be closely aligned with the objectives and constraints of large-scale forecast dissemination programs~\cite{afs_wmo_funding_news,wmo_call}. However, a major gap exists between current meteorologically focused benchmarks, which only assess the scientific validity of forecasts, and the adoption of such models for operational use and, ultimately, societal benefit. Designing decision-oriented benchmarks is complex, requiring consideration of AIWP development details, large-scale dissemination constraints, and an array of choices regarding variables, spatio-temporal scales, metrics, and selection criteria.

We introduced a framework that evaluates forecasts using a scientific, population-relevant index and a range of decision-oriented metrics in an operational setting against a relevant baseline and historical \textit{in-situ} observations. This framework bridges the gap between state-of-the-art meteorological benchmarking and decision-oriented benchmarking for large-scale adoption. Although this framework is also valuable for traditional NWP forecasts, AIWP models open new opportunities for LMICs to cost-effectively generate, tailor, and operationalize multiple models' forecasts for specific uses, as well as benchmark them for rare events (e.g., once-a-year monsoon onset) against out-of-sample local observations. Such benchmarking can also accelerate scientific advances by revealing societally relevant shortcomings, such as a model's high FAR and MR despite low MAE, which might be addressed by improving model architecture or loss function or applying bias correction and fine-tuning.

For Indian monsoon onset, across deterministic and probabilistic metrics over multiple test periods (Supplementary Table S2) and validated against local observations, nearly all models (AIWP and NWP) outperformed the climatological baseline for lead times up to $\sim 15$~days without any bias correction (Figures~\ref{fig2}-\ref{fig3} and Extended Data Figures 1-4). For longer lead times (16-30 days), the models generally had errors comparable to those of the baseline, but some ensemble forecasts (e.g., from IFS and NGCM) exhibited probabilistic skill out to $\sim 20$ days. These findings showed little sensitivity to the evaluation period, except for FuXi-S2S, suggesting the utility of using multiple out-of-sample and even in-sample periods for benchmarking to address the small test-sample-size issue for high-impact rare events.

It may seem surprising that forecasts of a local, rainfall-based onset index have skill, by several metrics, for lead times of 2-3 weeks, since forecasts of individual precipitation events exhibit low skill at such lead times~\cite{shrestha2013evaluation,surcel2015study}. However, monsoon onset is not a discrete precipitation event, but a transition from a dry, continental-scale dynamical regime to a wet one, with precipitation entering our decision-oriented index only after smoothing by a 5-day mean and local spatial averaging. {This onset index is, furthermore, what can be disseminated (e.g., via short text messages) as actionable information for farmers~\cite{burlig2024long}, demonstrating the utility of ``human-centered'' indices in benchmarking.}
Because the onset regime transition is influenced by a suite of incompletely understood couplings with remote ocean temperatures~\cite{goswami2006physical}, distant snow and ice cover~\cite{bamzai1999relation}, and land surface processes~\cite{chakraborty2023role}, there might be opportunities to further improve AIWP model skill for onset forecasting, as the current models have inadequate or no representation of these processes (see Methods). This sets a clear, high-impact target for future AIWP model development.

To further tailor the benchmarking for decision-makers' needs, we evaluated multiple deterministic and probabilistic metrics because no single metric captures forecast utility for all stakeholders. False alarms, for example, may have greater negative consequences for farmers than a large MAE if they cause farmers to plant too early and experience total crop loss. Probabilistic forecasts and associated metrics similarly enable users to adapt in the face of a particular risk;  e.g., a farmer might pivot to a different crop or even a different occupation if told there is a $90\%$ chance of a very late monsoon onset, yet be unwilling to make such a change if the probability is $50\%$. Showing that all the ensemble models are overconfident in probabilistic forecasts of onset (Figure~\ref{fig3}e), with none standing out as most skillful by all metrics, illustrates the need for post-processing and building multi-model, calibrated ensemble forecasts for dissemination\cite{aitken2026monsoonhybrid}.

The framework's operational value was demonstrated when it informed model selection for India's 2025 monsoon onset forecast dissemination, which employed the NGCM and AIFS models post-processed with historical statistics\cite{aitken2026monsoonhybrid}.  
Those two models were chosen based on several criteria: the ability to rapidly run models from available near real-time initial conditions (this excluded FuXi-S2S); skill at multiple weeks of lead time to best inform planting decisions (this excluded GraphCast); and both low {\it total} miss rates and low MAE (this excluded FuXi). The small sample size of the recent test period furthermore justified excluding AIWP models that could not be rapidly evaluated in other periods due to their computational cost (this excluded GenCast; see Methods). 
The spatial domain for dissemination of onset forecasts was chosen to avoid areas where the historical frequency of post-onset dry spells was large, and to exclude regions with high historical spatial variation of onset date within each grid-cell (Supplementary Figure S9).  After emerging from this selection process, operational forecasts from AIFS and NGCM were blended with historical rainfall statistics to create calibrated probabilistic forecasts of monsoon onset for 38 million farmers in the CMZ in the summer of 2025 \cite{aitken2026monsoonhybrid, vallangimonsoon25}. Yet even without post-processing, NGCM and AIFS individually predicted the anomalous evolution of early-season monsoon rainfall and a subsequent dry period 2-3 weeks in advance in 2025 (Figure~\ref{fig4}, Extended Data Figure 5, Extended Data Table 2, and Supplementary Table S6).

Operational, decision-oriented benchmarking frameworks are needed to support actionable early warning systems.
Moving from this decision-oriented benchmarking process to at-scale forecast dissemination requires post-processing to reduce model bias (including training-data bias) and calibrate predicted probabilities. It also requires designing messaging systems and message content that can appropriately guide decision-making in local contexts. Given recent progress in AIWP development\cite{pathak2022fourcastnet,bi2023accurate,lam2023learning,price2025probabilistic,yuval2024ngcmimerg,lang2024aifs,allen2025end,kossaifi2026demystifying}, emerging evidence for weather forecast benefits \cite{burlig2024long,hallegatte2012cost,rosenzweig2019assessing}, and decreasing cost of information transmission\cite{world2016world,fabregas2019realizing}, now is the time for interdisciplinary efforts to integrate these separate advances to improve livelihoods in LMICs through weather forecasts and climate adaptation tools.


\section*{Methods}

\subsection*{Global NWP and AIWP models and ground-truth data}

We evaluate the performance of multiple publicly available global AIWP models for forecasting local monsoon onset over India (Extended Data Table~1). These models (except for FuXi-S2S) are included in the benchmark because they can be transparently evaluated (i.e., their codes and weights are publicly available for inference to generate hindcasts), include precipitation as an output (necessary for a rainfall-based onset metric), and can be implemented operationally for the real-time dissemination of onset forecasts. These forecasts are compared with available hindcasts from a leading state-of-the-art NWP model: Integrated Forecasting System (IFS)\cite{ecmwf2023ifss2s} from the European Centre for Medium-Range Weather Forecasts (ECMWF). A brief description of each model and the specific version used in this study is provided below (see Extended Data Table~1 for a summary):

\paragraph{IFS (Cy48r1)~\cite{ecmwf2023ifss2s}:}
IFS is ECMWF's operational NWP model. It is a physics-based model, which in its subseasonal forecast configuration has a horizontal resolution of Tco319 ($\approx$ 32 km) and is coupled to the NEMO ocean model ($0.25^\circ$ horizontal resolution and 75 vertical levels) and an interactive sea-ice model. To ensure consistent evaluation, we utilize the 11-member hindcast dataset spanning 2004–2023 from a fixed model cycle (Cy48r1). 46-day hindcasts, available twice weekly (initialized on Mondays and Thursdays), are obtained from the S2S database~\cite{vitart2017subseasonal}.

\paragraph{AIFS~\cite{lang2024aifs}:}
The Artificial Intelligence Forecasting System (AIFS) is ECMWF's data-driven deterministic weather model. This model is based on a graph neural network architecture and a sliding window transformer processor. For this study, we use the first operational version (AIFS Single v1.0), which was pre-trained on ERA5 data from 1979 to 2022 and further fine-tuned on operational IFS analysis data from 2016 to 2022. AIFS does not use sea surface temperature, but predicts soil moisture and temperature (prognostic). 

\paragraph{FuXi~\cite{fuxi2023}:}
FuXi is a data-driven deterministic weather model that employs cube embedding and a transformer-based architecture. To optimize for both short and long lead times, FuXi uses a cascade architecture with pre-trained models, fine-tuned for 5-day forecast time windows of 0-5 days, 5-10 days, and 10-15 days. The model is trained on ERA5 data from 1979 to 2015, with 2016 and 2017 being used as validation years. FuXi does not use sea surface temperature or land information. 

\paragraph{FuXi-S2S~\cite{chen2024machine}:}
FuXi-S2S is a probabilistic data-driven subseasonal forecasting model that provides global daily mean forecasts for up to 42 days. The model is trained on ERA5 data from 1950 to 2016. Fuxi-S2S predicts sea surface temperature (prognostic) but does not include land information. Since the model code and model weights necessary for hindcast and forecast generation are not public, we could only use the publicly shared hindcasts available for 2002-2021. 

\paragraph{GraphCast~\cite{lam2023learning}: }
GraphCast is a data-driven deterministic weather model developed by Google DeepMind. It is based on a graph neural network architecture in an encoder-processor-decoder configuration. A few different trained versions of GraphCast exist; in this study, we use the 0.25$^\circ$ resolution, 37-pressure level version, which is trained on ERA5 data from 1979 to 2017. GraphCast does not use sea surface temperature or land information. 

\paragraph{GenCast~\cite{price2025probabilistic}:}
GenCast is a data-driven probabilistic weather model developed by Google DeepMind, which is based on a conditional denoising diffusion architecture. It learns the generative distribution of future atmospheric states, conditioned on past observations. For this study, we use the version of the model that is trained only on ERA5 data from 1979 to 2018. GenCast predicts sea surface temperature (prognostic) but does not use land information.

\paragraph{NeuralGCM (NGCM)~\cite{yuval2024ngcmimerg}:}
Google's NGCM is a hybrid model that integrates a physics-based differentiable dynamical core and a neural network-based single-column parameterization of subgrid-scale physical processes. A few different trained versions of NGCM exist; in this study, we use the version that is trained from 2001 to 2018 on ERA5 data, except for precipitation, for which IMERG is used. To generate forecasts, NGCM requires sea surface temperature and sea ice concentration as boundary forcings (not prognostic variables). To align with an operational use, we have used the climatological sea surface temperature and sea ice concentration.  

All of these AI models are trained to learn the atmosphere's ``fast dynamics'': they are optimized during training with a loss function that minimizes the mean error between a short {\it global} forecast (that could be from one 6-hourly time step to multiple time steps covering a few days) and the training data (often ERA5). In fact, the data-driven models mentioned above, more or less, have a loss function that looks like 
\begin{equation}
\label{eq:loss}
\mathcal{L}= \|x_{i}(t+\Delta t) - \mathcal{N}(x_{i}(t),b_i(t),\theta)\|.
\end{equation}
\noindent Here, $\mathcal{N}$ is a deep neural network, $x$ is the global 3D atmospheric (prognostic) state from ERA5 evolved from time $t$ to $t+\Delta t$, $b(t)$ represents forcings and boundary conditions (if any), $\theta$ are the trainable parameters of $\mathcal{N}$, and $i=1, 2 \dots N$ denotes the training samples. $\Delta t$ is often 6 hours or slightly longer (less than a day). $\| \cdot \|$ is a norm. During inference, the atmospheric state is forecasted using
\begin{equation}
\label{eq:inf}
x(t+\Delta t) = \mathcal{N}(x(t),b(t),\theta),
\end{equation}
in an autoregressive fashion, i.e., $x(t+2\Delta t) = \mathcal{N}(x(t+\Delta t),b(t+\Delta t),\theta)$ is predicted next and so on. While state-of-the-art AI weather models, such as those evaluated here, use different norms and variants of Eqs.~\eqref{eq:loss}-\eqref{eq:inf} (e.g., sometimes more terms or inputs are added), they overall use similar learning principles, and these equations are just presented to provide a high-level overview.

Note that all of the AI models use ERA5 precipitation for training, with the exception of NGCM, which was trained on satellite-based IMERG precipitation. The choice of training data source can influence a model’s precipitation forecast performance. ERA5 precipitation is known to exhibit systematic biases, particularly in the tropics \cite{lavers2022evaluation,hassler2021comparison}. In contrast, several studies have shown that IMERG provides reliable precipitation estimates over the Indian region \cite{dubey2021evaluation,prakash2018preliminary,sengupta2023assessing}, although notable uncertainties persist over orographic regions. During the onset phase (e.g., around June), both ERA5 and IMERG exhibit similar east–west patterns in mean accumulated rainfall bias over India (Supplementary Figure S8). Relative to the IMD rainfall (the ground-truth), ERA5 exhibits a dry bias over northwestern India and a wet bias over the southeastern region. While IMERG shows a similar spatial pattern, the biases are weaker in magnitude, including over the CMZ.

To generate hindcasts with the AIWP models, instead of selecting a specific event and initializing the models days in advance of the known onset, we follow a more operationally oriented approach, which, in particular, exposes the forecasts to false alarms and false negatives, which can have major impacts on the end user. Here, forecasts are initialized beginning in May of each year. Since the hindcasts for IFS \cite{ecmwf2023ifss2s} are available at a twice-weekly frequency, we have focused our evaluation of AIWP models on those exact dates for a direct comparison. All AIWP models, except for FuXi-S2S, are initialized using ERA5 data valid on the initialization date at 00 UTC. For FuXi-S2S, forecasts available on Hugging Face \cite{chen2024machine} are used. Note that NGCM requires boundary forcings for sea surface temperature and sea ice concentration (components of $b(t)$ in Eqs.~\eqref{eq:loss}-\eqref{eq:inf}), but these fields are not available in real-time for operational forecasting. We therefore prescribed ERA5 hourly day-of-year climatological fields from 1979 to 2017 as boundary conditions. Additionally, the operational NGCM and AIFS forecasts used for the 2025 monsoon season (Extended Data Figure 5) were initialized with NCEP GDAS (FNL) and ECMWF IFS analyses, respectively. Note that NCEP GDAS does not include the variables specific cloud liquid water content and specific cloud ice water content, nor does it provide all 37 pressure levels required by NGCM as initial conditions. These missing variables were derived from the cloud water mixing ratio using a temperature threshold of $-20^{\circ}\mathrm{C}$, and all fields were vertically interpolated using log-pressure interpolation from 33 levels (22 for cloud water mixing ratio) to the full set of 37 pressure levels. 

For ground-truth data, we use the daily $1^{\circ}\times1^{\circ}$ gridded rainfall from the IMD rain-gauge network \cite{rajeevan2006high}. This gridded dataset is based on rainfall records from a network of IMD observatory stations, hydro-meteorology observatories, agrometeorological observatories, and stations maintained by the state governments. This high-quality dataset, which is currently available from 1901 to 2024, has been used to study many aspects of monsoon dynamics and its variability \cite{mishra2021anthropogenic,roxy2017threefold,ghosh2012lack}, and is widely accepted as the best available ground-truth rainfall data over India. We choose $1^{\circ}$ over a higher available resolution of $0.25^{\circ}$ to prioritize temporal homogeneity and stable gauge support over nominal spatial resolution. Evidence suggests false trends and a step increase in gridded extremes around the mid-1970s, traced to evolving station availability\cite{lin2019if}. For the $1^{\circ}$-resolution product (1901–2024), IMD uses a fixed network of 1,380 rainfall stations \cite{rajeevan2008analysis}. 

Horizontal resolution varies across models (Extended Data Table 1). In order to compare the results across models, both IMD observations and models' forecasts are regridded to a common $4^{\circ}$ horizontal grid using conservative remapping. This reduces the spatial variability present at smaller scales. For models with a $0.25^{\circ}$ output resolution, a land–sea mask is first applied to retain only the land values prior to remapping. This step ensures that values over the ocean are excluded from the coarse $4^{\circ}$ grid, thereby preventing spurious contributions that could bias the evaluation. For models with coarser native resolutions, the influence of ocean values on the remapped grid was negligible, and therefore the land–sea mask was not applied. Moreover, the main focus of this study is to assess the onset skill of the models in the CMZ. At $4^{\circ}$ resolution, there are roughly 10 grid cells that cover the CMZ. 

The variability of onset dates within each $4^{\circ}$ grid using $1^{\circ}$ and $0.25^{\circ}$ data is also computed; for CMZ, most of this variability is observed in the northwestern boxes (Supplementary Figure S9). Such within-cell variability has important implications for decision-making about where to disseminate (see Discussion).

\subsection*{Periods of analysis}

Many AIWP models are trained on most of the post-1979 (satellite-era) years of ERA5 reanalysis , with most models either trained on data spanning from 1979 to 2018, and further fine-tuned on later years (see Extended Data Table 1). Consequently, only a few recent years remain as {\it out-of-sample} test data for forecast evaluation. For rare events such as weather extremes and monsoon onset over India (and other regions), which occur once a year, such a small test sample size can makes it challenging to obtain robust, reliable results. To mitigate this, we conduct the evaluation over 4 different periods, keeping in mind that each have specific advantages and shortcomings (Supplementary Table S2). These four different periods are examined separately: The \emph{Recent test period} (2019-2024), which is the out-of-sample period for most AIWP models; the \emph{Extended period} (1965-1978 and 2019-2024), which also contains pre-satellite out-of-sample years (except for FuXi-S2S, but the hindcasts from that period are unavailable anyway); \emph{All available years} (1965-2024), totaling 60 years including both training and testing years of AIWP models; a \emph{Common period} (2004-2021), created to assess the probabilistic performance of IFS and AIWP models on a longer shared period, which has no missing years for any model except for GenCast. Owing to GenCast's much higher computational cost of inference (around $\mathcal{O}(100)\times$ more time and node-hours compared to other AIWP models), hindcasts with this AIWP are produced only for the recent test and extended periods. Note that the forecasts from the pre-1979 period are susceptible to lower-quality initialization data (pre-satellite era). For IFS and FuXi-S2S, forecasts for only a subset of years are available (for all other models, we generated the forecasts).

\subsection*{Agriculturally Relevant Monsoon Onset Definition}

\subsubsection*{The Moron-Robertson onset index}

Many different definitions of monsoon onset exist in the literature based on large-scale and local-scale phenomena \cite{bombardi2020detection}. Among the different definitions, estimating the first wet spell is a key indicator of local rainfall-based monsoon onset. However, initial rainfall bursts can be followed by subsequent dry periods, making these first rainy bursts ``bogus'' (false) signals of the true start to the monsoon season. To distinguish ``true'' onset dates separately from ``false starts'',  Moron and Robertson \cite{moron2014interannual} established a local-scale onset criterion specifically for agricultural purposes. This definition uses a region-based minimum rainfall threshold sufficient for seed germination, defining monsoon onset with two criteria: (1) Following April 1, the first wet day ($\geq 1\mathrm{mm}$) of the first 5-day wet sequence that accumulates at least the amount of the local climatological 5-day wet spell (wet spell threshold; see Supplementary Figure S10a), and (2) The wet spell from (1) is not followed by any 10-day dry spell ($<5\ \mathrm{mm}$ in total) within the subsequent 30 days\cite{moron2014interannual}. 

From a forecasting perspective, predicting the Moron-Robertson-defined monsoon onset, i.e., satisfying both (1)-(2), is inherently challenging. This difficulty arises because verifying the second condition requires precipitation information well beyond the onset date. For example, an onset forecast issued 5 days in advance still necessitates accurate predictions extending up to approximately 35 days ahead to confirm the absence of a prolonged dry spell following the initial rains. Such extended lead times lie at the upper limits of subseasonal forecast skill, where uncertainties are considerably larger. 

\subsubsection*{Modified local-scale onset definition with MOK filter}

To enable the development of actionable onset forecasts several weeks in advance, we adopt a modified Moron-Robertson onset definition anchored to the \emph{median climatological onset date over Kerala}, which reduces the dependence on long-lead post-onset precipitation forecasts (criterion (2) above). IMD defines the onset of monsoon over Kerala (MOK), based on the fulfillment of specific criteria, including large-scale atmospheric circulation patterns, regional rainfall thresholds, and outgoing longwave radiation \cite{joseph2006summer,pai2009summer}. The MOK is typically followed by a northwestward advancement of the monsoonal rains. This typical inland northward progression of monsoon onset following MOK can be leveraged as an alternative indicator to minimize premature “false start” detections, without necessitating reliance on the full Moron-Robertson onset definition. Most of these false starts are observed in the northwestern semi-arid region of India and often occur in early May \cite{moron2014interannual}. Thus, we modify this onset definition such that the first wet spell after MOK is considered as the onset date over a region, regardless of whether or not it is followed by dry spells. However, in operational forecasting, the actual MOK date for a given year is unknown in advance; therefore, the climatological median MOK date is used as a reference date for filtering local onset. The median MOK date based on the past 124 years (1901-2024) is June 2nd (Supplementary Figure S11). Thus, a local onset at any given location is defined as the first day of the first wet spell observed after June 2nd. This modification leads to a reduction of $\sim$74\% in false onsets and places forecasts within the context of the nationally influential MOK event. Moreover, the climatological local onset date based on this modified definition clearly shows a typical northwestward progression of onset (Supplementary Figure S10b). This highlights the effectiveness of this definition in accurately capturing monsoon onset dynamics. However, a limitation of this definition is its inability to identify anomalously early onsets that may occur prior to the beginning of June. Earlier cut-off dates can be chosen based on the distribution of Supplementary Figure S11 to assess the robustness of the benchmarking results.

\subsubsection*{A large-scale circulation onset index}

In addition to local-scale onset based on precipitation, we also evaluate the models' ability to capture the large-scale circulation changes that occur during monsoon onset. An index that captures the change in the large-scale vertical wind shear and the related horizontal temperature gradient at the time of onset is the Webster-Yang index\cite{webster1992monsoon} (WYI). WYI is defined as:
\begin{equation}
    WYI = \overline{U}_{200hPa} - \overline{U}_{850hPa}
\end{equation}
where, $\overline{U}$ is the horizontally averaged zonal wind in the region of 40$^\circ$E-110$^\circ$E and 0-20$^\circ$N.
A large-scale onset based on this index is defined as the date when the seven-day moving average of WYI crosses its climatological value on June 2nd (median MOK date) \cite{chevuturi2021forecast}. 

\subsection*{Evaluation Metrics}

\subsubsection*{Deterministic metrics}
A systematic benchmarking algorithm was developed to evaluate the onset forecast skill of the models in an operational framework. For each forecast initialization date, fixed forecast windows were defined: the extended-range period (1–15 days), which represents the typical limit of weather forecasting, and the subseasonal period (16–30 days), which captures the subseasonal time scale. Forecasts were assessed for the occurrence of monsoon onset in each window. For example, for a forecast window of 15 days, for a model initialized on June 1st, the forecast would be evaluated for potential onset between June 2nd and June 16th. To satisfy the pentad-based threshold criteria for defining agriculturally relevant local onset, the algorithm requires a 19-day forecast time series following the initialization date.

If the model predicts onset within the forecast window, the model onset date is compared to the observed onset date (from IMD rainfall) for that year, and the absolute error is computed as the absolute difference between the two dates. This process is repeated for each initialization to compute the mean absolute error (MAE) for each year. Furthermore, if this error is within a predefined tolerance ($\pm3$ days for a 1-15 day forecast window and $\pm5$ for a 16-30 day window), the prediction is labeled a true positive (TP); otherwise, it is a false positive (FP). If no onset is predicted, the absence of an observed onset in the whole forecast (first 15 days and first 30 days for 1-15 day and 16-30 day forecast windows, respectively) is treated as a true negative (TN), whereas a missed observed onset is considered a false negative (FN). This procedure is applied iteratively across all grids until the nearest available initialization to the observed onset date is identified. This framework enables the computation of skill metrics such as MAE, FR, and MAR, in an operational-like setting, providing a robust assessment of model performance for monsoon onset forecasts. 

Three common deterministic metrics were chosen for benchmarking the models:
\begin{itemize}
    \item \textbf{Mean Absolute Error (MAE):} The absolute difference (in days) between the observed onset (based on IMD rainfall data) and the one obtained from the model forecast, averaged across all model initializations for a year. A lower MAE corresponds to better skill. For a given year \( k \), MAE is calculated as:
\begin{equation}
\text{MAE}_{k} = \frac{1}{n} \sum_{i=1}^{n} \left| y_i - y_{T} \right|
\end{equation}
    where:
\begin{itemize}
    \item \( n \) is the number of model initializations with onset forecast (up to the observed onset date),
    \item \( y_i \) is the onset forecasted by the model,
    \item \( y_{T} \) is the observed onset.
\end{itemize}
    
    \item \textbf{False Alarm Rate (FAR):} This metric estimates the rate at which the model incorrectly predicts onset within a forecast window when there is no actual onset observed for the same window. It is the ratio of false positives (FPs) to the sum of false positives and true negatives (TNs) for all model initializations:
    \begin{equation}
    \text{FAR} = \frac{\text{Total number of FPs}}{\text{Total number of FPs +  Total number of TNs}}
    \end{equation}
    
    \item \textbf{Miss Rate (MR):} This metric estimates the rate at which a model fails to predict monsoon onset when a true onset occurs within a forecast window. Quantitatively, it can be defined as the ratio of the number of false negatives (FNs) to the total number of actual onsets in the forecast window for all model initializations:
    \begin{equation}
    \text{MR} = \frac{\text{Total number of FNs}}{\text{Total number of true onsets in the forecast window}}
    \end{equation}
\end{itemize}

There is some subtlety in defining a false positive in this setting, since we are forecasting not only whether or not an event occurred but also \emph{when} it occurred. If a model forecasts onset, and this onset is within three days for a 15-day forecast window or within five days for a 30-day forecast window, we consider this forecast to be a true positive; if not, we consider it to be a false positive. This makes the FAR metric more sensitive: if a model predicts the onset tomorrow but the true onset is two weeks away, it counts against the model's FAR. However, this means that the total number of true onsets in the forecasting window is no longer the sum of the number of true positives and false negatives, since a true onset can be classified as a false positive. As a result, the denominator of the miss rate formula does not reduce to the sum of true positives and false negatives.

To obtain these metrics for a climatological baseline, a similar approach is implemented, where for a given initialization, the true onset is compared to the climatological onset date if it falls within the forecast window. Here, the climatological onset date at each grid cell is obtained as the average date of onset from 1901-2024 (Supplementary Figure S10b). This approach enables the climatological forecast to be evaluated in an operational framework, allowing the identification of misses and false alarms. In addition, we compute the MAE of a fixed climatological forecast (Supplementary Tables S1, S3–S5 and Supplementary Figure S7), obtained as the mean absolute difference between the climatological onset date and the observed onset across the analysis period at each grid cell. Furthermore, for the models, we also compute the total misses in the CMZ as the percentage of years in which no onset was detected for any initialization (from May 2 until the date of the true onset date) for the 10 grids in the CMZ (Supplementary Tables S1, S3–S5).

For deterministic evaluation of probabilistic models, the procedure described above is applied to each ensemble member. An onset forecast is recorded only if at least 50\% of members predict an onset for a given initialization. When this threshold is met, the mean onset date across all members with valid onset forecasts is used; otherwise, the case is classified as ‘no onset forecast’ for that initialization.

\subsubsection*{Probabilistic metrics}

We also evaluate the probabilistic skill of each ensemble-based model using common probabilistic metrics. One challenge for standard metrics is that a given ensemble member may not forecast onset at all (if, for example, the forecast predicts rain below the 5-day threshold for an entire forecast window). To account for this, we consolidated possible onset dates into short intervals, or ``bins''. For example, using a 5-day bin and a 15-day forecast window, we have bins corresponding to days 1-5, 6-10, 11-15, and after day 15. Ensemble members failing to forecast an onset in the 15-day window were interpreted as predicting the onset to occur after day 15. For climatology, an additional bin is used to capture the climatological probability of onset occurring before the initialization date. This bin is not used in computing the Brier Score (BS) or Area Under the Receiver Operating Characteristic Curve (AUC) to ensure a constant set of bins across forecasts. It is used for all models in computing the Ranked Probability Score, taking non-climatology forecasts as assigning a 0\% probability to this bin. 

 For each forecast $i$ and bin $j$ (i.e., days 1-5 after initialization, days 6-10 after initialization, etc.), from a given model, we obtained the probability $p_{ij}$ as the fraction of ensemble members (out of $w$ member) predicting onset in this bin. We define $Y_{ij} = 1$ if the onset in fact occurred in this bin, and 0 otherwise. Three standard probabilistic metrics were used:
\begin{itemize}
    \item \textbf{Fair Brier Score (BS)}: This metric, also called the mean squared error, is defined as: 
\begin{equation}
    \text{BS} =\dfrac{1}{n \cdot m} \sum_{i = 1}^n \sum_{j = 1}^m \left(\left(Y_{ij} - p_{ij} \right)^2 - \dfrac{p_{ij}(1-p_{ij})}{w-1}\right) 
\end{equation}

where $n$ is the number of forecasts (initializations), $m$ is the number of bins per forecast, and $w$ is the number of ensemble members. The Brier Score reflects both a forecast's calibration (whether assigned probabilities correctly capture long-run frequencies of events) and resolution (the extent to which the forecast ``resolves'' probabilities from the underlying climatological probability towards 0 or 1). Note that a lower Brier score corresponds to a better forecast, and that we use an adjustment according to Ferro et al. \cite{ferro2014fair} to account for differing ensemble sizes. 
\item \textbf{Fair Ranked Probability Score (RPS):} The RPS is a generalization of the Brier Score, which takes into account the distance between bins: for example, a prediction of onset in week 2 is ``closer'' to correctly predicting a week 3 onset than a prediction of onset in week 1. It is defined as: 
\begin{equation}
\text{RPS} = \dfrac{1}{n \cdot m} \sum_{i = 1}^n \sum_{k = 1}^m \left(\left(\sum_{j = 1}^k \left(Y_{ij} - p_{ij}\right) \right)^2 -\dfrac{ \left(\sum_{j = 1}^k p_{ij}\right)\left(1- \sum_{j = 1}^k p_{ij}\right)}{w-1}\right)    
\end{equation}
Note that we again use a standard adjustment to account for different ensemble sizes. 
 \item \textbf{Area Under the Receiver Operating Characteristic Curve (AUC): } The Receiver Operating Characteristic (ROC) curve is formed by using a probability threshold $t\in[0,1]$ for a binary classifier based on the model output and plotting the true positive rate against the false positive rate. These points define a curve, called the ROC curve, and the AUC is bounded between 0 and 1. The area under the ROC curve can be interpreted as the probability that, selecting a random forecast and bin in which onset occurred, and selecting a random forecast and bin in which onset did not occur, the model assigned a higher probability to the bin in which onset occurred than the bin in which onset did not occur. More formally, this can be computed as:
 \begin{equation}
 \text{AUC}  = \dfrac{\displaystyle\sum_{i,j,i',j'} Y_{ij}(1-Y_{i'j'}) \cdot 1[p_{ij} > p_{i'j'}]}{\displaystyle\left(\sum_{i,j} Y_{ij}\right)\left(\sum_{i,j} (1- Y_{ij})\right)}.    
 \end{equation}

Generally, improving the resolution of a forecast will increase its AUC; however, enhancing its calibration without improving its resolution will not. The AUC does not have a standard adjustment for ensemble size, so in general, increasing ensemble size will improve the AUC. Note that a higher AUC corresponds to a better forecast. 
\end{itemize}

Brier scores and ranked probability scores are then converted to \emph{skill} scores, defined as the percent improvement a given metric shows relative to climatology. Formally, for a metric $x$ and set of forecasts $i$, we define
\begin{equation}
\text{Skill}(x,i) = 1 - \dfrac{x_i}{x_{\text{climatology}}}  
\end{equation}

The climatological baseline is defined to be an ensemble forecast with one member for each year between 1901 and 2024. For example, if the true onset date for a given grid cell was June 12 in 1986, the 86th ensemble member would forecast June 12th for that grid cell, regardless of initialization date.  While this ensemble includes the year being predicted, the resulting error is small enough (less than 0.1\% of the overall scores) not to affect any of our analyses.

\section*{Acknowledgments}
This work grew out of a collaboration with India's Ministry of Agriculture and Farmers' Welfare.  We are grateful for fruitful discussions with Shreya Agrawal, Medha Deshpande, Genevieve Flaspohler, Subimal Ghosh, Ankur Gupta,  Neil Hausmann, Stephan Hoyer, Peter Huybers, Mrutyunjay Mohapatra, Vincent Moron, Raghu Murtugudde, Sivananda Pai, D.R. Pattanaik, Thara Prabhakaran, Satya Prakash,  V.S. Prasad, Suryachandra Rao, Muthalagu Ravichandran, Sakha Sanap, Sahadat Sarkar, K. K. Singh, Gokul Tamilselvam, and Janni Yuval. This work is partially supported by AIM for Scale, and by the University of Chicago's Human-centered Weather Forecasts Initiative (a program of the Institute for Climate and Sustainable Growth, ICSG), Development Innovation Lab, and AI for Climate Initiative (a joint program of the Data Science Institute, DSI, and ICSG). We thank the University of Chicago's DSI and Research Computing Center for computational resources.

\section*{Author contributions}
{PH, WB, AJ, and MK conceived and planned the study. PH, WB, and AJ designed the study and supervised the research. AM and HP generated the AIWP forecasts. RM, CA, AM, MG, TY, YQS, and HP analyzed data. All authors discussed and interpreted the results. RM, CA, KK, AJ, WB, and PH wrote the paper. AM, TY, YQS, and HP contributed to writing of the Methods section and Supplementary Materials. All authors reviewed and edited the manuscript.}

\section*{Data availability}
The ERA5 dataset used forinitializing the AIWP models is obtained from the Copernicus Climate Change Service (C3S) Climate Data Store. The real-time initial conditions used in the 2025 monsoon season forecasts are obtained from the NCEP GDAS/FNL product inventory (\url{https://www.nco.ncep.noaa.gov/pmb/products/gfs/#GDAS}) and the ECMWF IFS Open Data Portal (\url{https://data.ecmwf.int/forecasts/}). The IFS hindcasts are publicly available from the S2S database (\url{https://apps.ecmwf.int/datasets/data/s2s}). The FuXi-S2S hindcasts are obtained from Hugging Face (\url{https://huggingface.co/datasets/FudanFuXi/FuXi-S2S}). The IMD 1-degree gridded rainfall data are available from the IMD data portal (\url{https://www.imdpune.gov.in/cmpg/Griddata/Rainfall_1_NetCDF.html}). ECMWF's AIFS weights are obtained from \url{https://huggingface.co/ecmwf/aifs-single-1.0}. The model weights for Google DeepMind's GraphCast and GenCast are available at \url{https://github.com/google-deepmind/graphcast}. The weights for FuXi can be obtained from \url{https://zenodo.org/records/10401602}. Weights for Google Research's NeuralGCM (stochastic, IMERG precipitation) can be accessed at \url{https://neuralgcm.readthedocs.io/en/latest/checkpoints.html}. The regridded forecasts from all the models and the codes used for benchmarking in this study can be found at \url{https://doi.org/10.5281/zenodo.18407547} and \url{https://github.com/envfluids/monsoon-benchmark}.

\captionsetup[table]{name=Extended Data Table}
\captionsetup[figure]{name=Extended Data Figure}
\setcounter{figure}{0}  


\clearpage

\begin{table}[ht!]
\centering 
\setlength{\tabcolsep}{10pt}
\renewcommand{\arraystretch}{1.75}
\rowcolors{2}{gray!15}{white}
\begin{tabular}{c|c|c|c|c|c|c}
\hline
\textbf{Model} & 
\textbf{Model type} &
\makecell{\textbf{Type of forecast} \\ \textbf{(ensemble size)}} & 
\makecell{\textbf{Native} \\ \textbf{spatial}\\ \textbf{resolution}}  & 
\makecell{\textbf{Training} \\ \textbf{period}} & 
\makecell{\textbf{Fine-tuning} \\ \textbf{period}} & 
\makecell{\textbf{Generated} \\ \textbf{hindcast} \\ \textbf{period}} \\
\hline
\makecell{IFS \\ (Cy48r1)~\cite{ecmwf2023ifss2s}} & NWP  & Probabilistic (11) & $\approx32$ km & N/A & N/A & 2004-2023 \\

AIFS~\cite{lang2024aifs} & AIWP & Deterministic & $0.25^{\circ}$  & 1979-2022 & \Gape[0pt][2pt]{\makecell{2016-2022\\(IFS)}} & 1965-2024 \\

FuXi~\cite{fuxi2023} & AIWP & Deterministic & $0.25^{\circ}$  & 1979-2017 & No fine-tuning & 1965-2024 \\

GraphCast~\cite{lam2023learning} & AIWP & Deterministic & $0.25^{\circ}$ & 1979-2017 & No fine-tuning & 1965-2024 \\

GenCast~\cite{price2025probabilistic} & AIWP & Probabilistic (51) & $0.25^{\circ}$ & 1979-2018 & No fine-tuning & \makecell{1965-1978; \\2019-2024} \\

FuXi-S2S~\cite{chen2024machine} & AIWP & Probabilistic (51) & $1.5^{\circ}$  & 1950-2016 & No fine-tuning & 2002-2021 \\

\makecell{NeuralGCM \\(NGCM)~\cite{yuval2024ngcmimerg}} & AIWP & Probabilistic (51) & $2.8^{\circ}$ & 2001-2018 & No fine-tuning & 1965-2024 \\

\hline
\end{tabular}
\caption{\textbf{Summary of models benchmarked for monsoon onset forecasts}. See the Methods for more details about each model. All hindcasts, except for IFS and Fuxi-S2S, have been generated as part of this project using open-source models initialized with ERA5 initial conditions (note that Extended Data Figure 5 and Supplementary Table S6 show results with operational analysis initializations). GenCast's hindcasts have been limited to two periods due to the computational cost of inference. IFS and FuXi-S2S data are obtained from public repositories (see Data Availability).}\label{raw-models}
\end{table}

\begin{table}[h!]
\centering
\begin{tabular}{llccc}
\hline
\textbf{Forecast Type} & \textbf{Metric} & \textbf{Climatology} & \textbf{AIFS} & \textbf{NGCM} \\
\midrule\addlinespace[6pt]

\multirow{3}{*}{\begin{tabular}{l}Deterministic \\ (1--15 day)\end{tabular}}
& MAE(days) & 6 & 2.3 & 2.5 \\
& FAR(\%) & 17.5 & 3.8 & 8.2 \\
& MR(\%)  & 39 & 18.5 & 17 \\

\addlinespace[6pt]\midrule\addlinespace[6pt]

\multirow{3}{*}{\begin{tabular}{l}Deterministic \\ (16--30 day)\end{tabular}}
& MAE(days) & 6 & 6.8 & 4.9 \\
& FAR(\%) & 28.6 & 34 & 29.9 \\
& MR(\%)& 17.6  & 23.9 & 11.8 \\

\addlinespace[6pt]\midrule\addlinespace[6pt]

\multirow{3}{*}{\begin{tabular}{l}Probabilistic \\ (1--15 day)\end{tabular}}
& AUC & 0.94 & - & 0.96 \\
& BSS(\%) & 0 & - & 18 \\
& RPSS(\%) & 0 & - & 19 \\

\addlinespace[6pt]\midrule\addlinespace[6pt]
\multirow{3}{*}{\begin{tabular}{l}Probabilistic \\ (1--30 day)\end{tabular}}
& AUC & 0.898 & - & 0.895 \\
& BSS(\%) & 0 & - & 4.7 \\
& RPSS(\%) & 0 & - & -1.3 \\

\addlinespace[6pt]
\bottomrule
\end{tabular}

\caption{\textbf{Forecast skill of AI weather models over the CMZ compared to climatology for the 2025 summer monsoon onset.} Forecast skill of AIFS (deterministic) and NGCM (probabilistic) is compared against climatology using the same metrics as the ones that used for benchmarking. Similar to the hindcast analysis, these scores are computed across twice-weekly initializations from 2 May 2025 until the observed onset date for each grid point. As before, deterministic scores are averaged and probabilistic scores are aggregated over the ten grid points covering the CMZ. Preliminary IMD data~\cite{nandi2024imdlib} for 2025 is used as the ground truth.
}

\end{table}

\begin{figure}[hbt!]
  \centering
  \includegraphics[width=\linewidth]{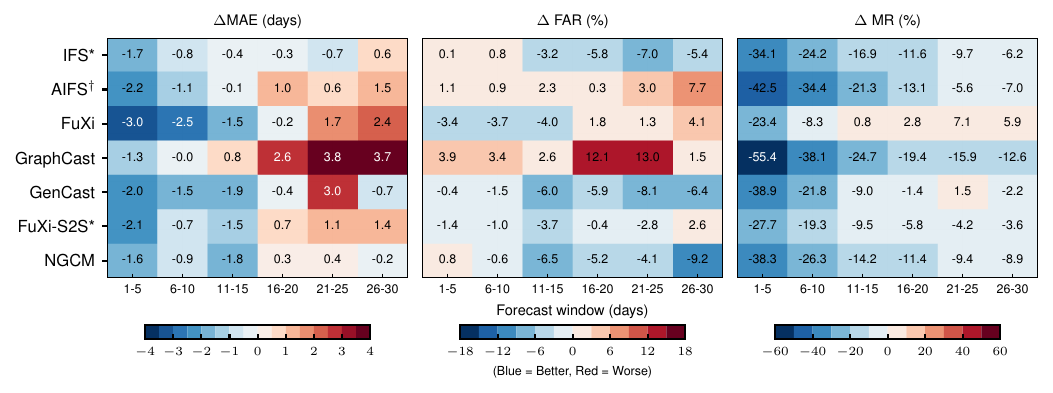}
  \caption{\textbf{Models' deterministic skill compared to climatology in 5-day forecast windows}. Difference of CMZ-averaged MAE, FAR, and MR of onset forecasts and climatology within 5-day forecast windows for the recent test period 2019-2024. Negative values (blue) mean better than climatology; positive values (red) mean worse than climatology. For computing FAR, the classification of an onset forecast being true positive or false positive is based on tolerance of $\pm2$, $\pm2$, $\pm3$, $\pm3$, $\pm5$, and $\pm5$ days for 1-5, 6-10, 11-15, 16-20, 21-25, and 26-30 day forecast windows, respectively. $*$ and $\dagger$ have the same meaning as in Figure~\ref{fig2}.}
  \label{extfig2}
\end{figure}

\begin{figure}[hbt!]
  \centering
  \includegraphics[width=0.8\linewidth]{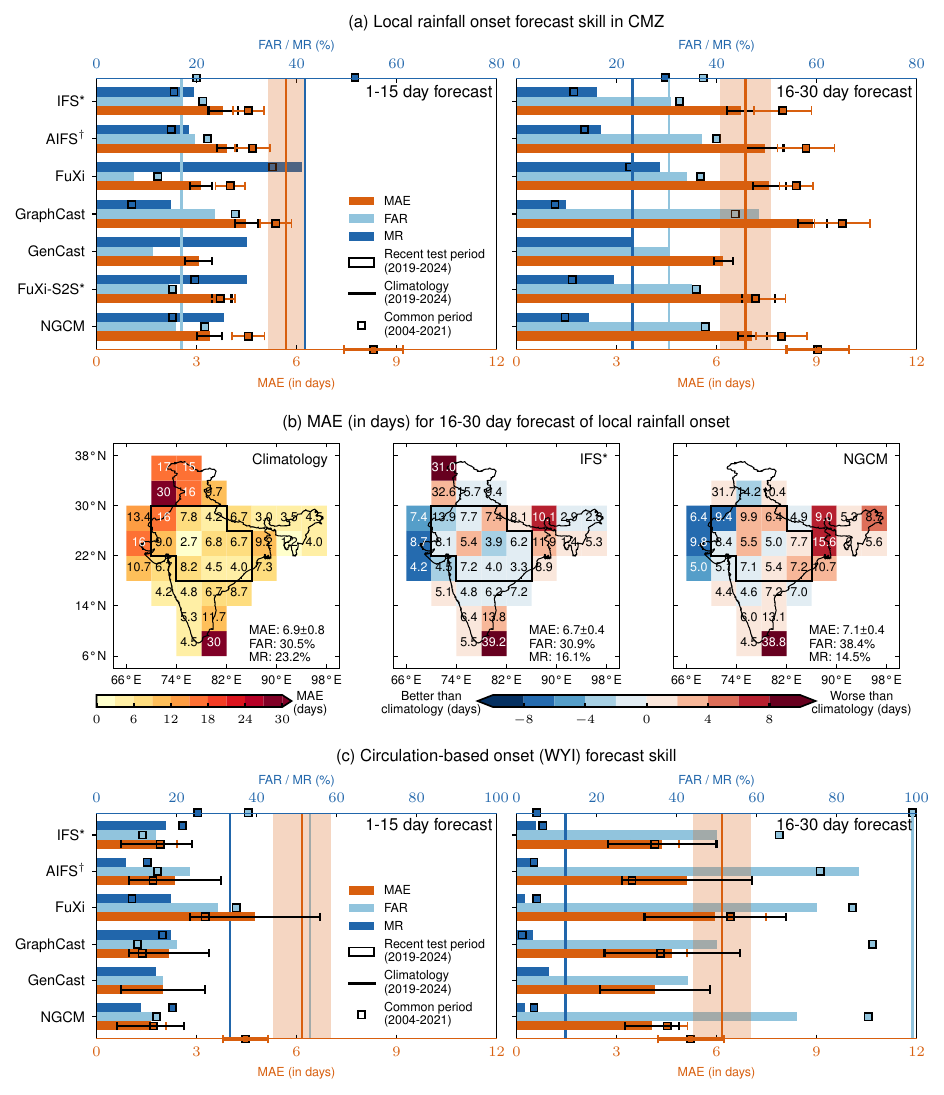}
  \caption{\textbf{Deterministic forecast skill of the models in the medium-range (1-15 day) and subseasonal (16-30 day) timescales.} a) and c): Same as Figure 2 but with the square markers showing the common period (20042021), which includes AI models' training years. b): Same as Figure 2 but for the subseasonal time scale in the recent test period (2019-2024).
  }
  \label{extfig1}
\end{figure}

\begin{figure}[hbt!]
  \centering
  \includegraphics[width=\linewidth]{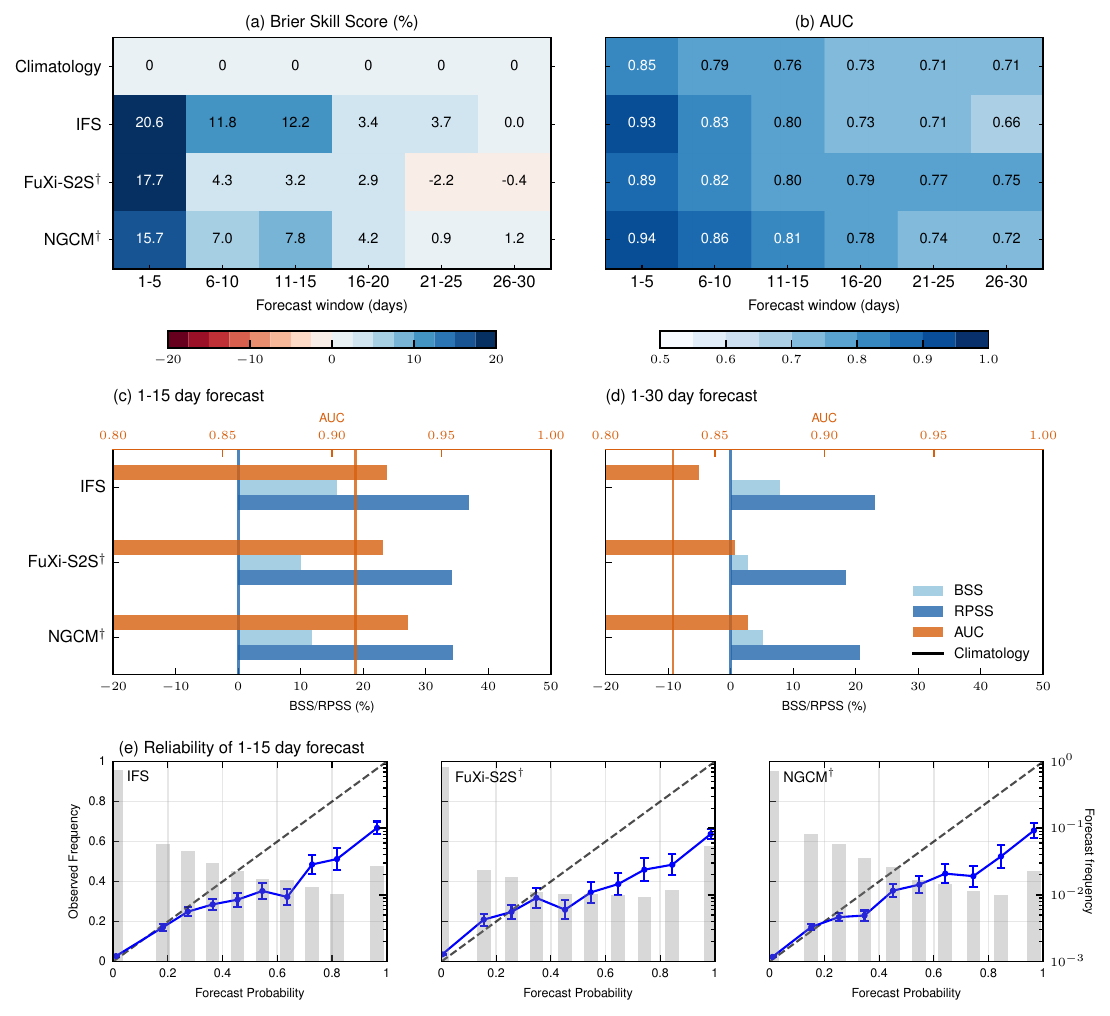}
  \caption{\textbf{Probabilistic performance of the models for the common period of 2004-2021.} This is the same as Figure~\ref{fig3} but the common period for available forecasts. (a) Brier Skill Score (BSS) and (b) AUC scores for 5-day binned forecast in the 10 grids of the CMZ. Aggregated probabilistic scores for (c) 15- and (d) 30- day forecast for this period. Climatological baseline for respective metrics in this period is presented as vertical lines. (e) Reliability diagram for onset forecast from the probabilistic models for a 1-15 day forecast window. Models with $\dagger$ next to them have some of their training years within this period. GenCast's forecasts were not produced for this period due to computational constraints.}
  \label{extfig3}
\end{figure}

\begin{figure}[hbt!]
  \centering
  \includegraphics[width=\linewidth]{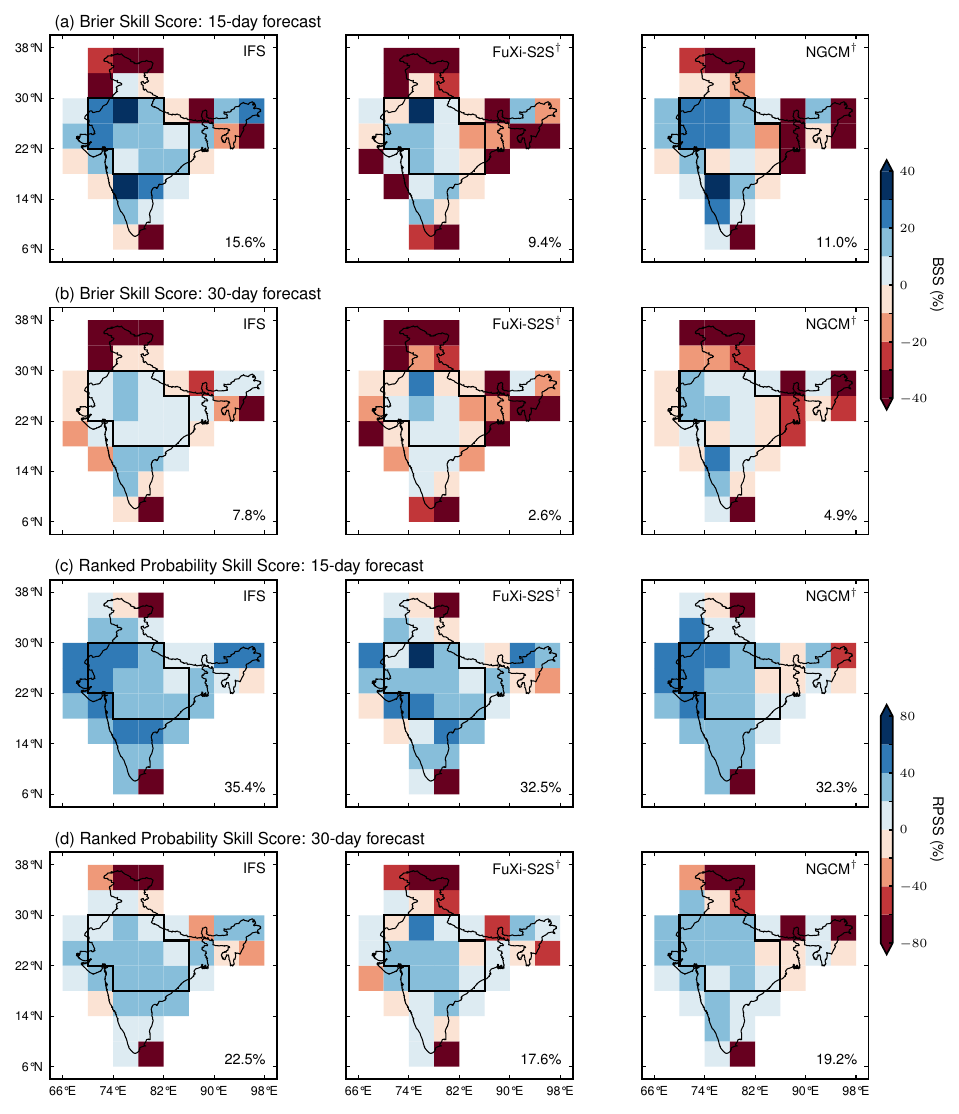}
  \caption{\textbf{Regional probabilistic skill of ensemble models for the common period of 2004-2021.} Note that this period covers part of the training period of the AIWP (denoted by $\dagger$). Panels (a)–(b) show the Brier Skill Score (BSS) and panels (c)–(d) show the Ranked Probability Skill Score (RPSS) for 15-day and 30-day forecasts, evaluated at each grid point. Positive skill values (blue) indicate performance better than climatology (and the higher, the better), whereas negative values (red) indicate performance worse than climatology. The CMZ is outlined by the bold black boundary, with the regional mean skill reported in the lower-right corner of each panel.
}
  \label{extfig4}
\end{figure}

\begin{figure}[htb!]
  \centering
\includegraphics[width=\linewidth]{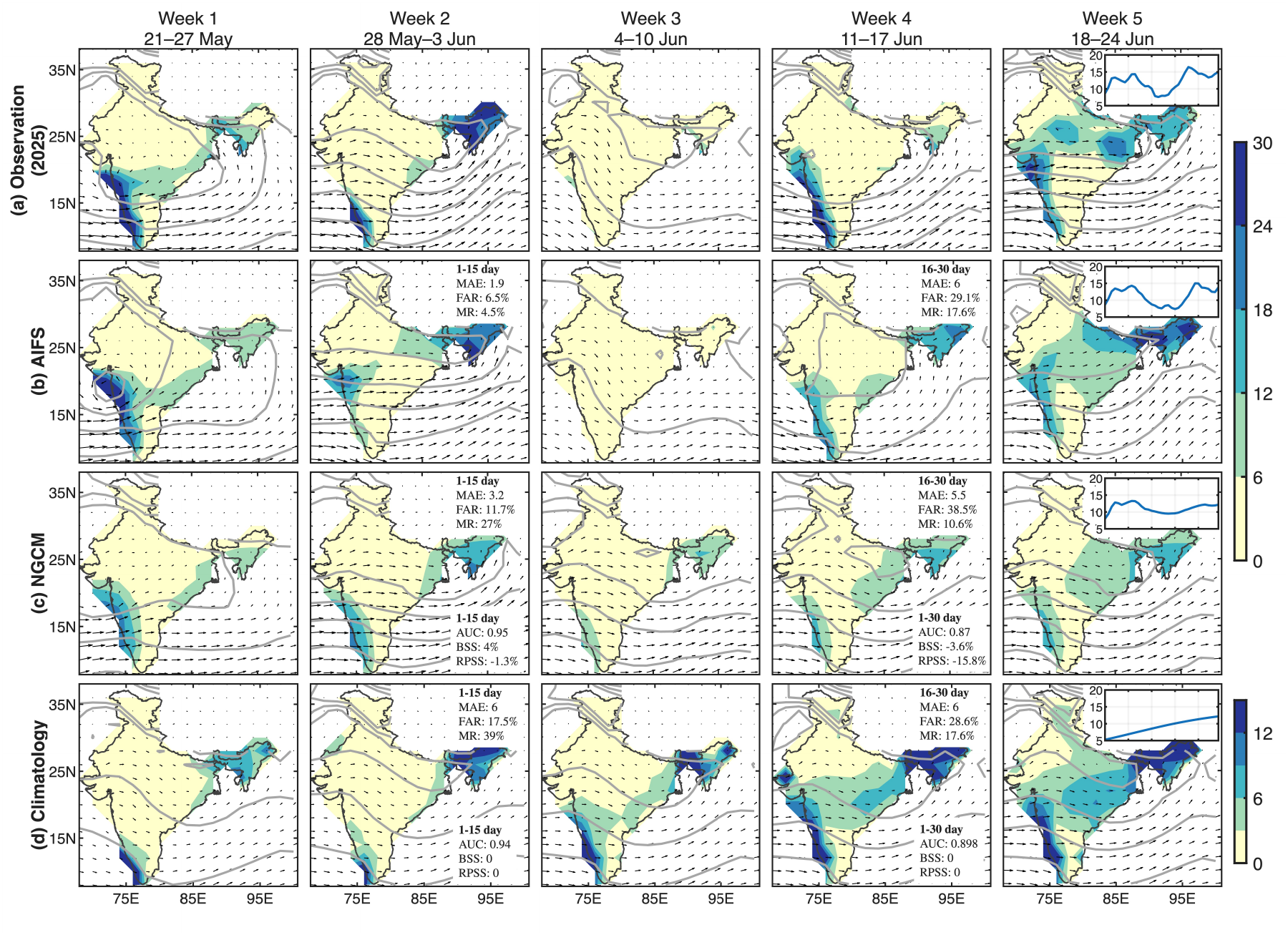}
\caption{\textbf{Monsoon 2025 forecasts initialized with real-time operational analysis data.} Same as Figure \ref{fig4} but for AIFS and NGCM forecasts initialized with real-time operational analysis from IFS and NCEP GDAS (FNL), respectively. See Supplementary Table S6 for more details about the skill scores shown on the panels.
}
  \label{extfig5}
\end{figure}
\bibliography{benchmarking_paper}


\end{document}